\documentclass{article}

    \usepackage[preprint, nonatbib]{neurips_2025}

\usepackage[utf8]{inputenc} % allow utf-8 input
\usepackage[T1]{fontenc}    % use 8-bit T1 fonts
\usepackage{hyperref}       % hyperlinks
\usepackage{url}            % simple URL typesetting
\usepackage{booktabs}       % professional-quality tables
\usepackage{amsfonts}       % blackboard math symbols
\usepackage{nicefrac}       % compact symbols for 1/2, etc.
\usepackage{microtype}      % microtypography
\usepackage{xcolor}         % colors
\usepackage{amsmath}
\usepackage{graphicx}
\usepackage{multirow}
\usepackage{wrapfig}
\usepackage{float}
\usepackage{adjustbox}
\usepackage{subcaption}

\title{GDI-Bench: A Benchmark for General Document Intelligence with Vision and Reasoning Decoupling}

\usepackage{booktabs}
\usepackage{pifont}
\usepackage{tcolorbox}
\newcommand{\cmark}{\ding{51}}  % check mark
\newcommand{\xmark}{\ding{55}}  % cross mark
\author{%
  Siqi~Li$^{1,2}$
  \quad
  Yufan~Shen$^{1}$ 
  \quad
  Xiangnan~Chen$^{1,2}$
  \quad
  Jiayi~Chen$^{3}$
  \quad 
  Hengwei~Ju$^{1,4}$ \\
  \textbf{Haodong~Duan$^{1}$}
  \quad
  \textbf{Song~Mao$^{1}$}
  \quad
  \textbf{Hongbin~Zhou$^{1}$}
  \quad
  \textbf{Bo~Zhang$^{1}$} 
  \quad
  \textbf{Bin~Fu$^{1}$}
  \quad
  \textbf{Pinlong~Cai$^{1}$} \\
  \textbf{Licheng~Wen$^{1}$}
  \quad
  \textbf{Botian~Shi$^{1, 5}$}
  \quad
  \textbf{Yong Liu$^{2,\dag}$}
  \quad
  \textbf{Xinyu~Cai$^{1,*, \dag}$}
  \quad
  \textbf{Yu Qiao$^{1}$} \\
  $^1$ Shanghai Artificial Intelligence Laboratory
  $^2$ Zhejiang University \\
  $^3$ School of Science and Engineering, The Chinese University of Hong Kong \\
  $^4$ Fudan University 
  $^5$ Shanghai Innovation Institute
}

\begin{document}

\maketitle

\renewcommand{\thefootnote}{\relax}
\footnotetext{* project leader, $\dag$ corresponding author. }
\renewcommand{\thefootnote}{\arabic{footnote}}

\begin{abstract}

The rapid advancement of multimodal large language models (MLLMs) has profoundly impacted the document domain, creating a wide array of application scenarios.
This progress highlights the need for a comprehensive benchmark to evaluate these models' capabilities across various document-specific tasks.
However, existing benchmarks often fail to locate specific model weaknesses or guide systematic improvements. 
To bridge this gap, we introduce a General Document Intelligence Benchmark (GDI-Bench), featuring 2.3k images across 9 key scenarios and 19 document-specific tasks.
By decoupling visual complexity and reasoning complexity, the GDI-Bench structures graded tasks that allow performance assessment by difficulty, aiding in model weakness identification and optimization guidance.
We evaluate various open-source and closed-source models on GDI-Bench, conducting decoupled analyses in the visual and reasoning domains, revealing their strengths and weaknesses.
To address the diverse tasks and domains in the GDI-Bench, we propose a GDI-Model that mitigates catastrophic forgetting during the supervised fine-tuning (SFT) process through an intelligence-preserving training strategy, thereby reinforcing the inherent weaknesses of the base model. 
Our model achieves state-of-the-art performance on previous benchmarks and the GDI-Bench.
Both our benchmark and models are or will be open-sourced on \url{https://huggingface.co/GDIBench}.

\end{abstract}

\section{Introduction}

Rapid progress of large language models (LLMs) \cite{liu2024deepseek, guo2025deepseek} has placed multimodal large language models (MLLMs) \cite{Qwen-VL,gpt4,team2023gemini} as a foundation of artificial intelligence, advancing document intelligence to a general stage. 
Cross-domain and multi-scale document understanding and extraction challenges are increasingly critical in real-world applications.
With the rise of MLLMs, a series of more complex benchmarks have emerged to provide comprehensive frameworks for document understanding tasks~\cite{fu2024mmecomprehensiveevaluationbenchmark,wu2024autohallusion,Guan_2024_CVPR,li2023seed2,li2024seed2plus,zhao2024tabpedia,feng2024docpedia}.
However, given that document understanding involves multiple modalities, errors in MLLM outputs may arise from inaccurate visual recognition, limited language organization, or both. Consequently, a decoupled evaluation of MLLMs’ document processing abilities is essential.

\begin{figure}[t]
    \centering
    \includegraphics[width=0.99\linewidth]{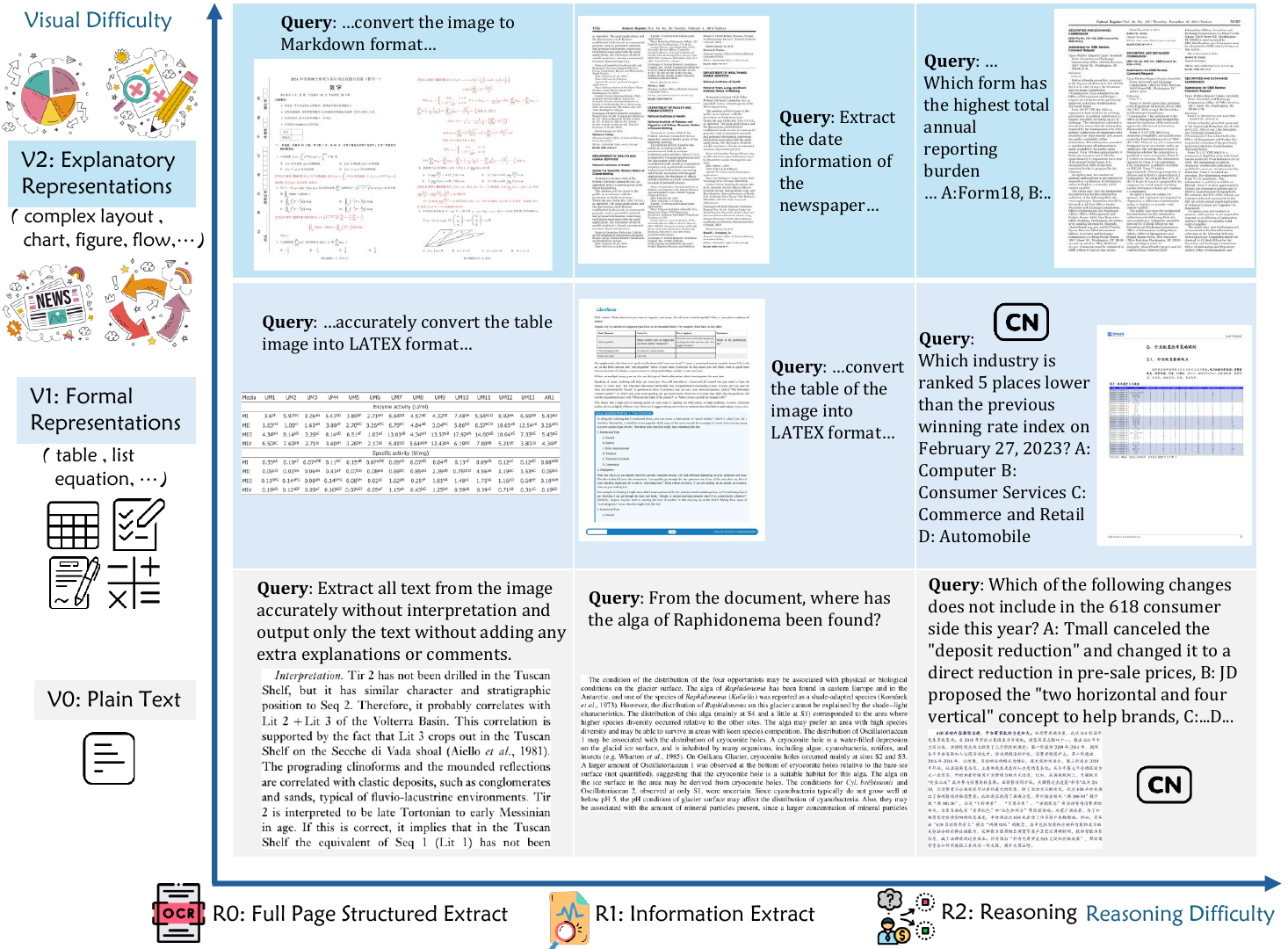}
    \caption{Overview of GDI-Bench. The benchmark decouples document understanding complexity into visual complexity (V0-V2) and reasoning complexity (R0-R2) dimensions, creating a comprehensive evaluation framework for assessing MLLMs' capabilities across various document types and reasoning tasks. Queries marked with a “CN” tag originate in Chinese and have been translated into English using Google Translate.}\label{figure:GDI_overall}
\end{figure}

To address these challenges, we propose the General Document Intelligence Benchmark (GDI-Bench), illustrated in Fig.~\ref{figure:GDI_overall}, which aims to perform a decoupled evaluation of model performance on document tasks, thereby contributing to the identification of the model's weaknesses.
The GDI-Bench introduces three key improvements over existing benchmarks: (1) developing a cross-domain, multi-task benchmarks to ensure task diversity and fine-grained difficulty levels; (2) introducing complexity decoupling, dividing multimodal document understanding into visual complexity and reasoning complexity, and for the first time, establishing a difficulty grading mechanism; (3) supporting the evaluation of MLLMs, OCR+LLM-level systems, and document parsing tools, offering comprehensive guidance for practical application solutions.

\begin{figure}[h]
    \centering
    \includegraphics[width=1\linewidth]{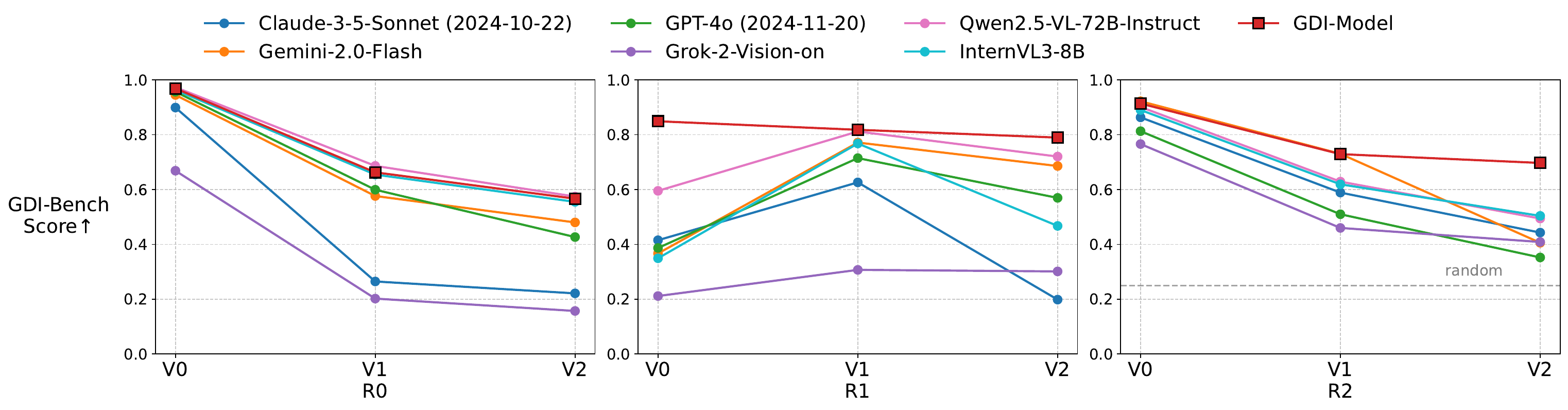}
    \caption{Performance of various open-source and closed-source models on GDI-Bench at different levels of reasoning complexity. The GDI-Model is fine-tuned based on the InternVL3-8B model.}\label{gdi-bench-eval}
\end{figure}

We evaluate 2 open-source and 4 closed-source large-scale models using the GDI-Bench and find their performance to be suboptimal, as shown in Fig. \ref{gdi-bench-eval}.
For instance, while the InternVL3-8B model performs well in the R0 domain but shows significant performance degradation in the R1 and R2 domains
To address the model’s weaknesses and further validate GDI-Bench’s capacity for weakness localization, we constructed supervised fine-tuning (SFT) on the InternVL3-8B model to explore whether data-driven approaches could improve its performance.
However, we observe that supervised fine-tuning tends to cause the issue of catastrophic forgetting \cite{kirkpatrick2017overcoming}.
To address this problem, we propose the Layer-wise Adaptive Freezing-Tuning (LW-AFT) method, which alleviates the impact of catastrophic forgetting and enhances the model's cross-domain and cross-task capabilities.
Specifically, during the SFT process, LW-AFT freezes most of the parameters, with only a small subset of domain-sensitive parameters participating in gradient updates.

Our contributions are summarized as follows.
\begin{itemize}
    \item We propose the GDI-Bench, a benchmark covering a broad range of document-related tasks. By decoupling complexity and grading difficulty, it helps the model identify its weaknesses and guides subsequent optimization.
    \item We propose Layer-wise Adaptive Freeze-Tuning, a training method that effectively alleviates catastrophic forgetting in document task SFT by parameter-freezing, improving performance on specific tasks while maintaining generalization capabilities.
    \item We propose a GDI-Model that achieves state-of-the-art (SOTA) performance on multiple document domain benchmarks as well as the GDI-Bench, demonstrating high generalization capabilities suitable for real-world applications.
\end{itemize}

\section{Related Works}

\paragraph{Document Benchmark}
Early benchmarks in document understanding primarily focused on single-domain tasks. For example, DocVQA\cite{qi-etal-2022-dureadervis} tackled industrial document QA, VisualMRC\cite{VisualMRC2021} focused on web-based comprehension, and ChartQA\cite{masry-etal-2022-chartqa} specialized in chart-based QA. While these benchmarks made significant contributions, they lacked difficulty grading and cross-domain generalization, due to the limited capabilities of models at the time. With the advent of MLLMs, more complex benchmarks emerged~\cite{fu2024mmecomprehensiveevaluationbenchmark,wu2024autohallusion,Guan_2024_CVPR,li2023seed2,li2024seed2plus,zhao2024tabpedia,feng2024docpedia}, offering a more comprehensive framework for document understanding. OCRBenchV2\cite{fu2024ocrbenchv2improvedbenchmark} expanded OCR evaluation with various subtasks, while Fox\cite{liu2024focus} improved document parsing through ROI-based methods. OmniDocBench\cite{ouyang2024omnidocbenchbenchmarkingdiversepdf} extended evaluation to cross-modal tasks, covering a wider range of parsing challenges. Despite these advancements, these benchmarks still lack explicit difficulty grading, limiting their ability to assess performance across tasks of varying complexity.

\paragraph{Document Understanding Model}
Optical Character Recognition (OCR) has long been a fundamental task in computer vision.
Existing OCR models can be broadly categorized into component-based and end-to-end approaches. 
Component-based methods~\cite{wang2024mineru,blecher2023nougat,paddleocrv2_du2021pp} adopt a modular pipeline that assembles multiple expert-designed components such as layout analysis~\cite{zhong2019publaynet}, text detection~\cite{tian2016detecting,liao2017textboxes,zhou2017east,liu2019curved}, region extraction, and content recognition~\cite{lecun1998gradient_mnist,graves2006connectionist,li2023trocr}.
In contrast, end-to-end OCR models~\cite{wei2409general}, especially those driven by MLLMs\cite{llava,Qwen-VL,wei2023vary,ye2023mplug,chen2024far_intervl1.5,liu2024textmonkey,liu2024focus_fox}, aim to unify perception and reasoning within a single architecture. 
Most MLLM-based OCR systems utilize CLIP~\cite{radford2021learning} as the vision backbone, while coupling it with a language model to jointly process visual and textual information in a unified framework.
Recent works~\cite{chen2024far_intervl1.5,liu2024textmonkey,ye2023ureader} adopt sliding-window strategies that partition the image into patches to cope with long and high-resolution inputs like PDFs.

\paragraph{Continual Learning of LLMs}
Continual learning in large language models (LLMs) faces the core challenge of catastrophic forgetting. Existing approaches mainly use two strategies: data replay ~\cite{sun2020lamol,kanwatchara-etal-2021-rational} and parameter freezing \cite{dou2024loramoe,razdaibiedina2023progressive,liu2023teamwork}.
Data replay mitigates forgetting by revisiting prior task samples during new task training. LAMOL\cite{sun2020lamol}, for example, generates pseudo-samples to avoid storage-based replay, while other methods like experience replay and interleaved task training integrate old data into the loop.
Parameter freezing protects existing knowledge by limiting parameter updates, using techniques such as LoRA\cite{hu2022lora}, adapters~\cite{adapter}, or prompt tokens~\cite{wang2022preserving}. Regularization methods like Elastic Weight Consolidation (EWC)~\cite{kirkpatrick2017overcoming} preserve key weights via importance constraints, while newer methods like Task Vector~\cite{ilharco2023editing} and Gradient Projection~\cite{saha2021gradient} operate at the gradient or representation level to reduce task interference and enhance generalization.
In contrast to these approaches, which typically rely on limited pre-training datasets, our method offers a more efficient and comprehensive alternative.

\section{Benchmark}

\subsection{Complexity Decoupling}

As illustrated in Fig.~\ref{figure:GDI_overall}, a difficulty-decoupled evaluation protocol named General Document Intelligence Benchmark (GDI-Bench) is introduced to comprehensively assess MLLMs' capabilities in visual information comprehension and reasoning. Distinct from VisualSimpleQA\cite{wang2025visualsimpleqabenchmarkdecoupledevaluation}'s structural decomposition of fact-seeking question answering tasks into visual recognition and knowledge dimensions through MLLM model architecture analysis, the proposed framework characterizes task complexity from fundamental difficulty perspectives by decoupling it into visual complexity and reasoning complexity. Notably, knowledge popularity is explicitly subsumed within the reasoning complexity dimension in this formulation.

\subsubsection{Visual Complexity}

\begin{wrapfigure}[17]{r}{2.8in} 
    \vspace{-2.5em}
\begin{minipage}{2.8in}
    \centering
    \includegraphics[width=1\linewidth]{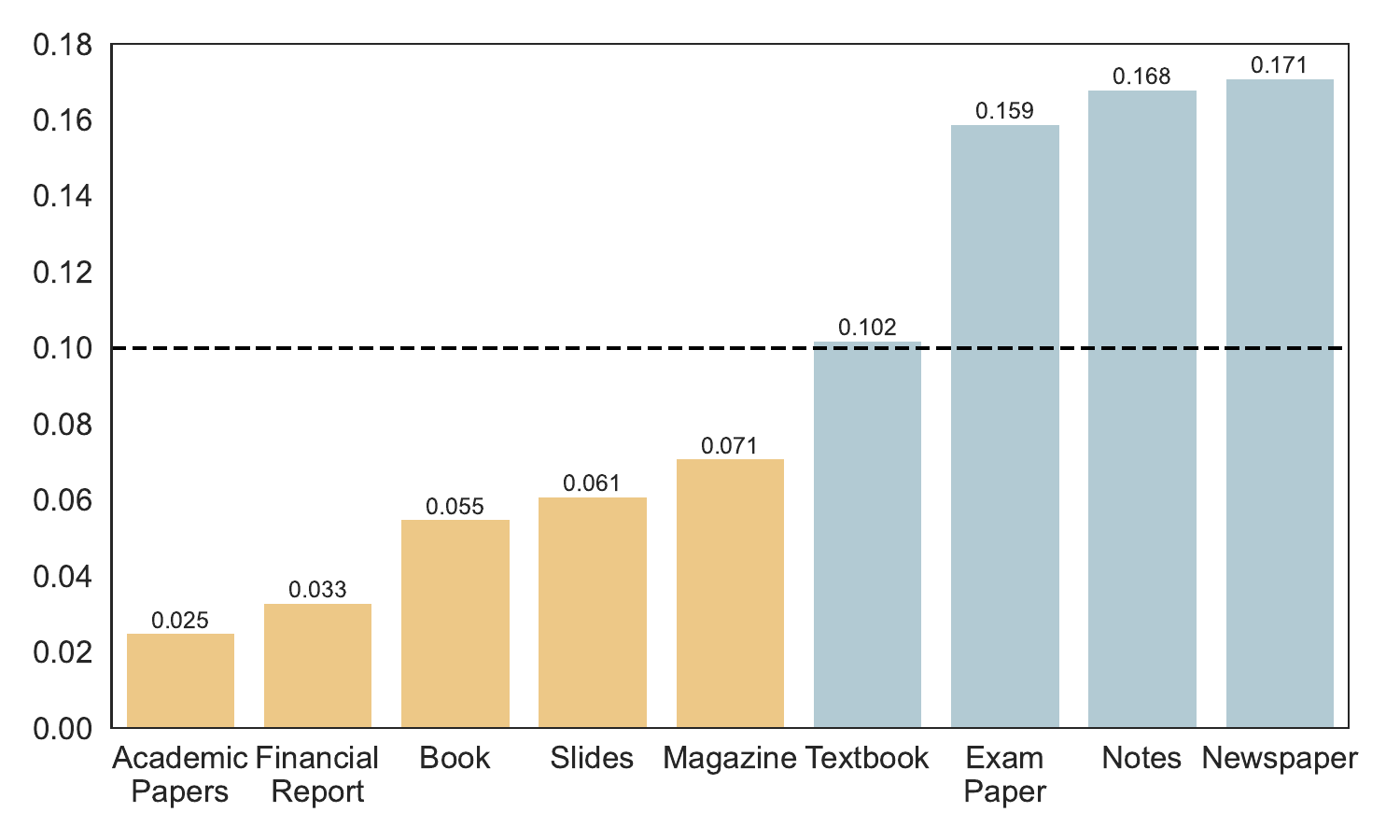}
    \caption{Distribution of visual complexity scores across nine document categories as reported in the OmniDocBench. The SOTA performance results from the benchmark show a substantial gap between document types, providing empirical justification for our visual complexity taxonomy. }\label{fig:omnidoc_performance}
\end{minipage}
\end{wrapfigure}

The visual complexity dimension is operationalized through a hierarchical categorization of document images into three levels: V0 (plain text), V1 (formal representations), and V2 (explanatory representations). V0 exclusively contains unstructured textual elements such as headings and paragraphs. Multimodal tasks on V0 documents typically achieve satisfactory performance via OCR-LLM pipeline architectures.
As demonstrated in Fig.~\ref{fig:omnidoc_performance}, a systematic analysis of the OmniDocBench \cite{ouyang2024omnidocbenchbenchmarkingdiversepdf} benchmark reveals statistically significant performance gaps. This benchmark covers nine document categories. Current pipeline tools and MLLMs show notably worse performance on textbook, exam paper, notes, and newspaper types compared to other forms. These performance drops directly correlate with complex features like multi-column layouts and graphical elements. Based on these observations, we designate these four challenging categories as V2 and classify the remaining five as V1. This creates a data-driven taxonomy for visual complexity characterization.

\subsubsection{Reasoning Complexity}

The reasoning complexity characterization is formulated through a behavior-driven taxonomy. Three distinct levels are defined to progressively evaluate document understanding capabilities. It is specifically categorized into R0: Full Page Structured Extract, R1: Information Extract, and R2: Reasoning.
Among these, the R0 task in V0 is more similar to OCR, and requires the ability to understand layout information such as tables, formulas, and complex page structures in V1 and V2 type images.
For R1 tasks, the model needs not only to recognize visual content but also to understand the task itself in order to accurately extract relevant information. This often involves interpreting lines, tables, labels, and other visual elements present in V1 and V2 images.
R2 tasks go a step further by requiring deeper reasoning capabilities, including the comprehension of logical inference and the ability to synthesize information across different modalities or layouts.

\subsection{Annotation Process}

\subsubsection{Data Source}
For the data collection phase of the GDI-Bench, documents are primarily sourced from OmniDocBench \cite{ouyang2024omnidocbenchbenchmarkingdiversepdf}, a dataset containing document images from 9 domains. GDI-Bench is further supplemented with an in-house collection of various document types, including exam papers, reports, newspapers, and so on. This multi-source integration results in a well-rounded and representative dataset that captures the multifaceted nature of real-world document understanding tasks.
\subsubsection{Data Construction}
\begin{wrapfigure}[12]{r}{3.2in} 
\begin{minipage}{3.2in}
\includegraphics[width=1\linewidth]{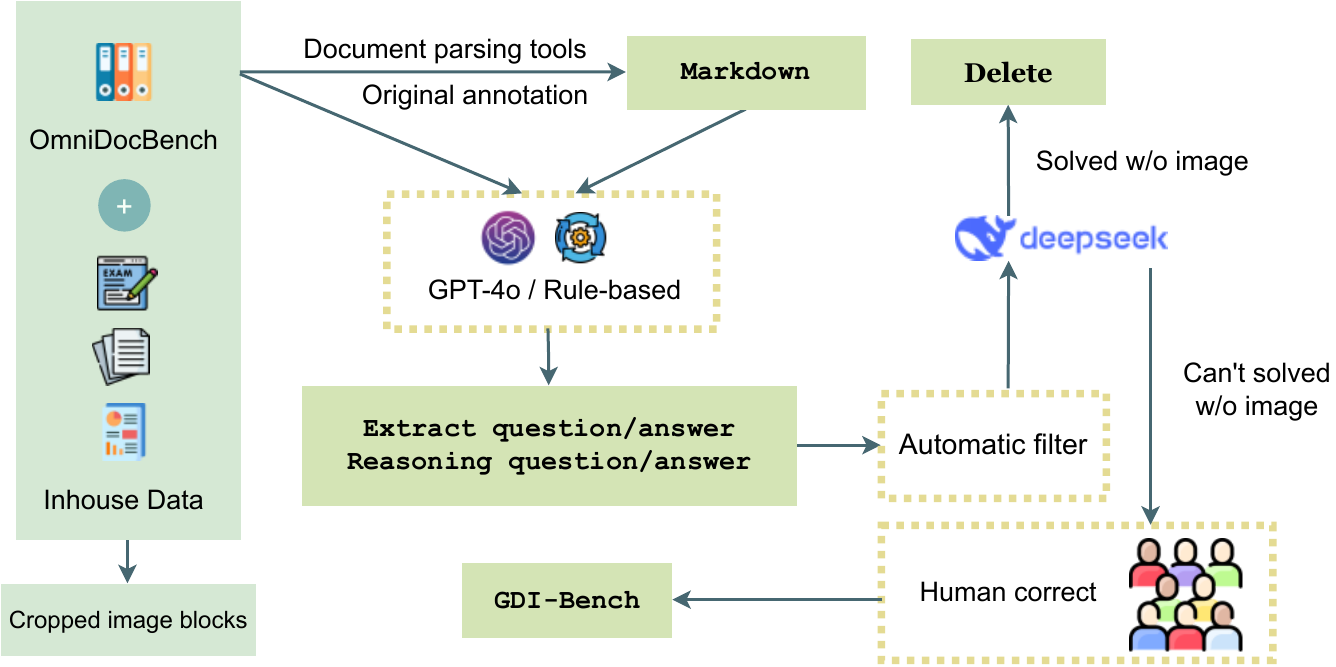}
    \caption{Annotation Process of GDI-Bench.}\label{fig:gdibench_generate}
\end{minipage}
\end{wrapfigure}

As shown in Fig.~\ref{fig:gdibench_generate}, the data construction process begins with cropping single-layout sub-images from OmniDocBench\cite{ouyang2024omnidocbenchbenchmarkingdiversepdf} and inhouse documents to form the V0 raw image set. Based on end-to-end edit distance scores from SOTA models and pipeline tools on OmniDocBench, domains with scores above 0.142 are categorized as V2, indicating high visual complexity. The remaining samples are assigned to V1.

For task construction, R0 tasks are generated using the original annotations or by synthesizing Markdown representations through MinerU \cite{wang2024mineru}. To create R1 and R2 tasks, both the Markdown and corresponding images are input into GPT-4o to generate extractive and reasoning question-answer pairs. In addition, rule-based tasks are designed manually. These include extracting specific questions from exam papers, retrieving values enclosed in colored boxes, and so on.

To ensure high data quality, all synthetic cases are first evaluated by models. Cases that can be answered correctly by DeepSeek-R1\cite{guo2025deepseek} without requiring document input or that are flagged as low quality are filtered out. Finally, a team of PhD-level annotators reviews and verifies the remaining instances to ensure accuracy and correctness in the final benchmark.

The final GDI-Bench contains a total of 3,660 test cases, distributed across different visual complexity levels (V0, V1, V2) and task types (R0, R1, R2), as summarized in Table~\ref{table:rv_statistics}. 

\begin{figure}[ht]
  \centering
\begin{minipage}[t]{0.38\textwidth}
  \centering
  \captionof{table}{Complexity Distribution in the GDI Benchmark.}
  \renewcommand{\arraystretch}{1.3}
  \label{table:rv_statistics}
  \resizebox{\textwidth}{!}{
  \begin{tabular}{ccccc}
    \toprule
    & \textbf{V0} & \textbf{V1} & \textbf{V2} & \textbf{Total} \\ \midrule
    \textbf{R0} & 377 & 521 & 450 & 1,348 \\
    \textbf{R1} & 273 & 498 & 288 & 1,059 \\
    \textbf{R2} & 298 & 641 & 314 & 1,253 \\ \hline
    \textbf{Total} & 948 & 1,660 & 1,052 & 3,660 \\ \bottomrule
  \end{tabular}}
\end{minipage}
  \hfill
  \begin{minipage}[t]{0.59\textwidth}
    \centering
    \captionof{table}{Comparison of GDI-Bench with Other Document Understanding Benchmarks.}
    \label{tab:generalization}
    \resizebox{\textwidth}{!}{
    \begin{tabular}{lcccc}
        \toprule
        \textbf{Benchmark} & Scenario & Task & Image &\textbf{Difficulty Grading} \\
        \midrule
        Seed-bench-2-plus &~8&1&0.6k& \xmark \\
        MMTab-eval &1&9&23k&\xmark \\
        MMC &1&9&1.7k& \xmark \\
        OCRBenchV2   &31 & 23&9.5k& \xmark  \\
        Fox        &2&9&0.7k  & \xmark  \\
        \midrule
        \textbf{GDIBench} &9&19&2.3k&\textbf{\cmark}  \\
        \bottomrule
    \end{tabular}}
    \end{minipage}
\end{figure}

\subsection{Evaluation Metrics}

We adopt different evaluation metrics depending on the task type.  
For reasoning tasks, most questions are designed in a single-choice format to ensure standardized evaluation.  
For non-choice tasks, such as document parsing and extraction (mainly R0 and R1), where answers typically appear verbatim in the source document, we use the \emph{Normalized Edit Distance} (NED)~\cite{levenshtein1966binary}, but compute the score as \(1 - \text{NED}\) so that higher scores correspond to better performance. Formally, the case score is:
\begin{equation}
\text{case}_i =
\begin{cases}
1 - \text{NED}(\text{prediction}_i, \text{ground truth}_i) & \text{for non-choice tasks} \\
1 & \text{for choice-based tasks and correct answer} \\
0 & \text{for choice-based tasks and incorrect answer}
\end{cases}
\label{eq:case_score}
\end{equation}

The overall GDI-Bench Score is the average over all evaluation cases:
\begin{equation}
\text{GDI-Bench Score} = \frac{1}{N} \sum_{i=1}^{N} \text{case}_i
\label{eq:gdi_score}
\end{equation}
where \(N\) is the total number of cases. A higher score indicates better overall performance.

\subsection{Comparison with Existing Document Intelligence Benchmarks}

As shown in Table~\ref{tab:generalization}, most existing document intelligence benchmarks are limited in either scenario coverage, task diversity, or data scale. 
In contrast, the GDI-Bench offers a balanced combination of 9 diverse document scenarios and 19 representative tasks, based on a curated set of 2.3k images. Notably, it introduces difficulty grading, a feature absent in all other compared benchmarks. 

\section{Methodology}\label{method}

\begin{figure*}[ht]
  \begin{minipage}[t]{0.39\textwidth}
  \centering
    \includegraphics[height=4cm]{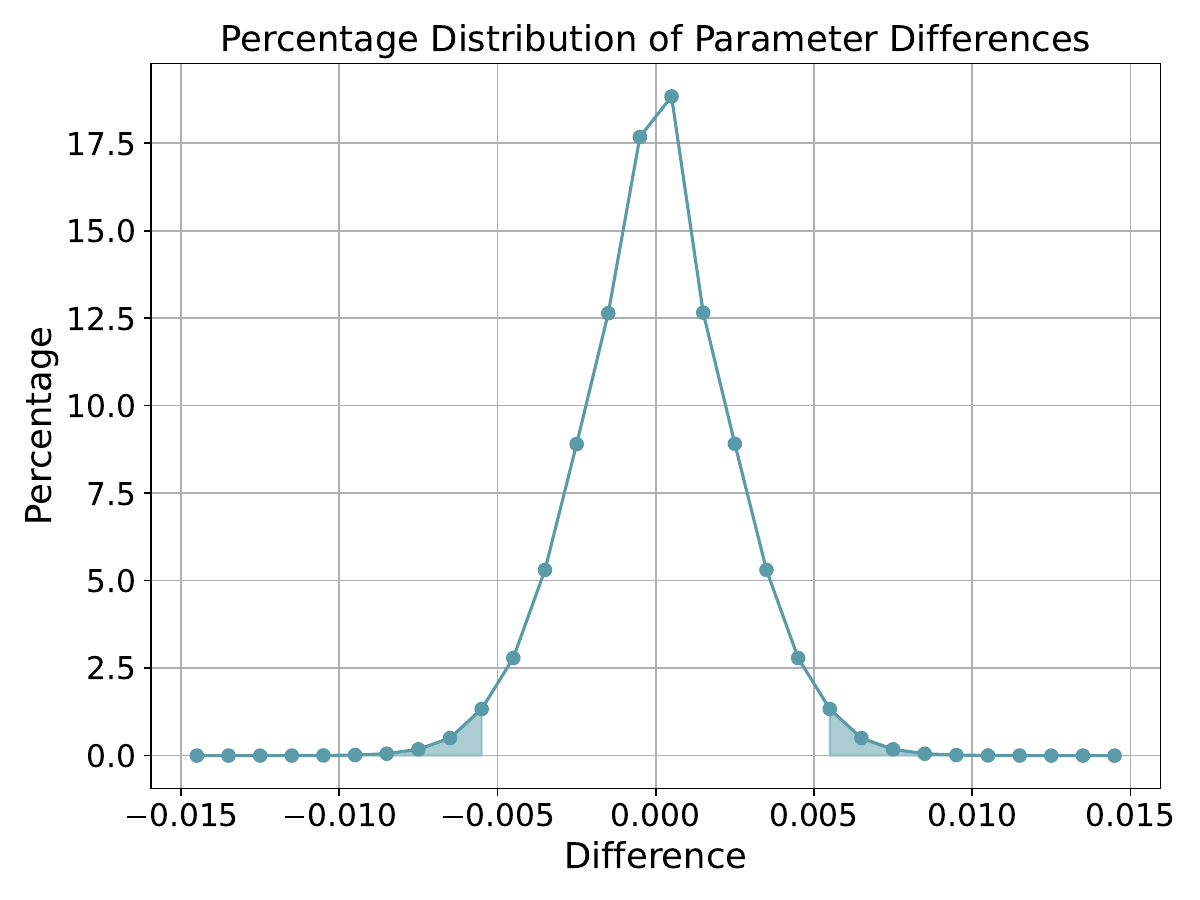}
    \caption{Distribution of parameter differences before and after full-parameter SFT.}
    \label{8bpara}
  \end{minipage}
  \hfill
  \begin{minipage}[t]{0.59\textwidth}
  \centering
    \includegraphics[height=4cm]{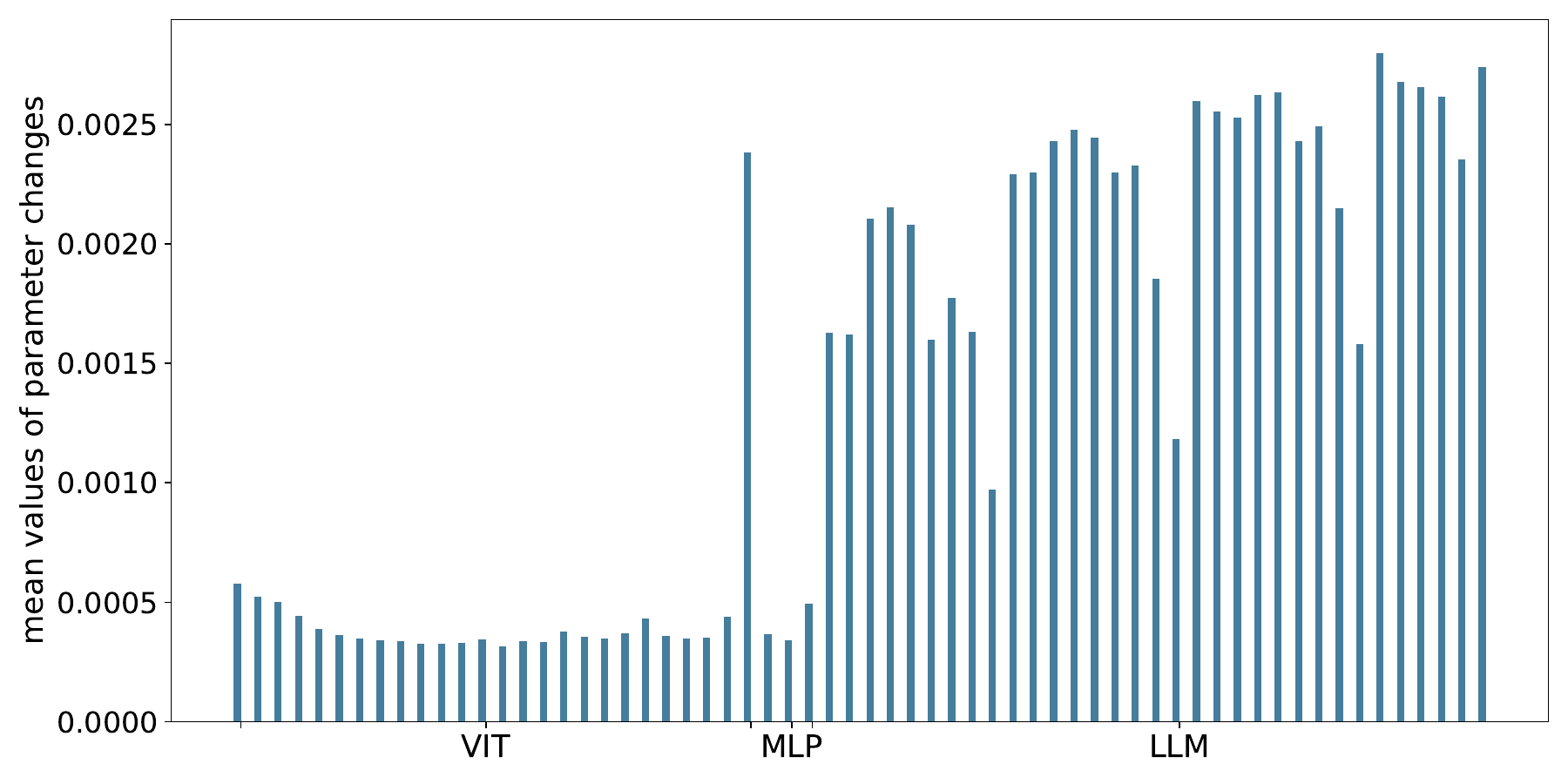}
    \caption{The mean values of parameter changes per layer before and after full-parameter SFT using 10\% of the training set.}
    \label{8blayer}
  \end{minipage}
\end{figure*}

We evaluated the InternVL3-8B model on GDI-Bench, conducting a detailed weakness analysis. As shown in Fig.~\ref{gdi-bench-eval}, the model performs well on R0 but drops significantly on R1 and R2, highlighting a large performance gap. 
To validate the ability of GDI-Bench to identify model weaknesses, we aimed to address these gaps in R1 and R2.
For this purpose, we constructed a multi-source training dataset that is structurally aligned with the tasks in R1 and R2, but sourced from entirely different data domains, detailed in App. \ref{trainingset}.
We performed SFT on the InternVL3-8B model using the constructed training set. However, subsequent evaluations revealed significant catastrophic forgetting, where the model lost important knowledge acquired during pretraining. As shown in Table~\ref{gdi-bench}, the Full-Parameter Fine-Tuning Model exhibited degradation in the R0 domain, where it had previously performed well.

To investigate the phenomenon, we comprehensively analyze parameter update dynamics in models subjected to full-parameter supervised fine-tuning.
Our experimental results reveal significant sparsity in learned parameter adaptations: over 95\% of parameters exhibit minimal changes, while a critical sparse subset (5\%) undergoes substantial modification (>0.005), as depicted in Fig.~\ref{8bpara}. 
Inspired by the Lottery Ticket Hypothesis \cite{frankle2018lottery}, we integrate this discovery with the hypothesis and formulate our theoretical proposition:
For any pre-trained MLLM with parameters $\Theta = \{\theta_i\}_{i=1}^n$, there exists a sparse task-salient sub-network $\hat{\Theta} = \{\hat{\theta}_j\}_{j=1}^m$ ($m < n$) and a domain-specific transformation $\psi$, such that $\psi(f_{\hat{\Theta}}(x))$ achieves performance comparable to the full model $f_{\Theta}(x)$ on the target domain $\mathcal{D}$. 
\begin{equation}\label{Lottery LLM Hypothesis}
\forall x \in \mathcal{D},\ \mathcal{P}(f_{\Theta}(x)) \approx \mathcal{P}(\psi(f_{\hat{\Theta}}(x)))
\end{equation}
where $\mathcal{P}(\cdot)$ is the metric for evaluating the performance of the model, and the subnetwork $\hat{\Theta}$ consists of parameters exhibiting high update sensitivity during SFT. 
Given that only a small subset of parameters exhibit significant task-specific adaptation, we propose performing gradient updates solely on these parameters during SFT, while freezing the majority of remaining parameters to maximally preserve the model's generalization capabilities and foundational knowledge.

\begin{figure}[t]
    \centering
    \includegraphics[width=1\linewidth]{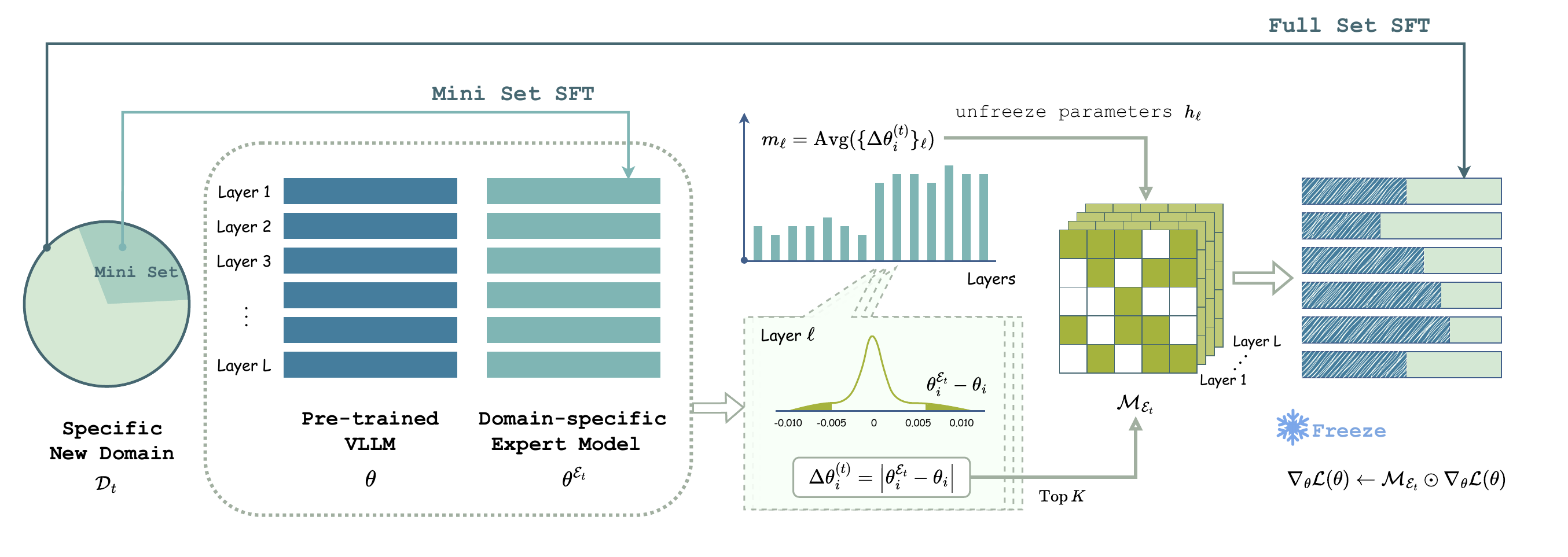}
    \caption{Overview of the Layer‑wise Adaptive Freeze‑Tuning method.}\label{overview}
\end{figure}

We formalize this approach as Layer‑wise Adaptive Freeze‑Tuning (LW‑AFT), which identifies and updates only critical parameters through layer-wise sensitivity analysis.
The overview of our method is shown in Fig. \ref{overview}.
For a specific new domain $\mathcal{D}_t$, the expert model (denoted as $\mathcal{E}_t$) is fine-tuned based on the pre-trained parameters $\Theta$ using only a subset $\alpha$ ($\alpha<1$) of the full dataset. For each domain-specific expert model, the element-wise absolute parameter variation is computed as:
\begin{equation}
    \Delta \theta^{(t)}_i = \left| \theta^{\mathcal{E}_t}_i - \theta_i \right|.
\end{equation}
We conduct a comprehensive layer-wise analysis of parameter updates $\Delta\theta^{(t)}_i$ before and after SFT.
Fig.~\ref{8blayer} illustrates the average magnitude of updates per layer, revealing distinct trends between the vision and language components.  
Specifically, the language layers exhibit a higher degree of parameter modification compared to the visual layers. This observation aligns with the problem localization we performed using GDI-Bench, where the InternVL3-8B model shows a performance degradation in reasoning dimensions.
Intuitively, we assume that the layers undergoing substantial parameter updates are critical for domain adaptation, as their dynamics closely align with task-specific knowledge transfer.
Conversely, layers with minimal updates appear less specialized to domain-specific features, and preserving their stability supports cross-domain generalization. 

Therefore, we implement a layer-adaptive freezing parameter allocation strategy. The model is divided into $L$ architectural layers with parameter counts $W = \{w_1,...,w_L\} \subseteq \mathbb{R}$.
Given a global budget of $H$ unfrozen parameters, we employ a proportional allocation mechanism where each layer $\ell$ is allocated $h_\ell$ unfrozen parameters, computed as follows:
\begin{equation}\label{hl}
    h_{\ell} = \frac{m_\ell \cdot w_\ell}{\sum_{i=1}^L m_i\cdot w_i}\cdot H
\end{equation}
where $m_\ell = \text{Avg}(\{\Delta \theta_j^{(t)}\}_\ell)$ is the average absolute change of the $\ell$-th layer.

After determining the number of unfrozen parameters for each layer $ \ell $, we first sort the absolute changes $ \{\Delta \theta_j^{(t)}\}_\ell$ in descending order, and then select only the top $ h_\ell$ parameters for further updates. This selection is performed via the function:
\begin{equation}
\phi_{h_\ell}: \mathbb{R}^{w_\ell} \to \{0,1\}^{w_\ell},\quad  
\phi_{h_\ell}\left(\{\Delta\theta^{(t)}_j\}_\ell\right) = 
\begin{cases} 
1, & \text{if } j \in \underset{k}{\mathrm{arg\,TopK}}\left(\{\Delta\theta^{(t)}_k\}_\ell, h_\ell\right) \\ 
0, & \text{otherwise}
\end{cases}
\end{equation}  
The binary masks for the entire network of expert $\mathcal{E}_t$ can be formulated as $\mathcal{M}_{\mathcal{E}_t} =  \left\{\phi_{h_\ell}\bigl(\{\Delta \theta_j^{(t)}\}_\ell \bigr)\right\}_{\ell=1}^L$. During the subsequent fine-tuning on a target domain, gradient updates are dynamically masked to exclusively modify parameters corresponding to $\mathcal{M}_{\mathcal{E}_t}$ through:
\begin{equation}
    \nabla_{\theta} \mathcal{L}(\theta) \leftarrow \mathcal{M}_{\mathcal{E}_t} \odot \nabla_{\theta} \mathcal{L}(\theta),
\end{equation}
where $\odot$ denotes the Hadamard product. 
By maintaining the stability of the base network dynamics and avoiding large parameter shifts, LW-AFT effectively mitigates catastrophic forgetting while adapting efficiently to downstream tasks. 
This approach not only reduces the number of parameters that need to be trained but also preserves global generalization capabilities, without relying on experience replay or other complex fine-tuning techniques. 
With the LW-AFT method, we are able to retain the outstanding performance of the InternVL3-8B model in the R0 domain, while improving its performance in the R1 and R2 domains, as shown in Table \ref{gdi-bench}. This further demonstrates the ability of GDI-Bench to pinpoint model weaknesses.

\section{Experiments}\label{experiments}
The experiments in this section consist of two parts.
In Section \ref{LW‑AFT}, we discuss the impact of our proposed Layer-wise Adaptive Freeze-Tuning (LW-AFT) method. In Section \ref{GDIbench}, we perform a comprehensive benchmark evaluation on the GDI-Bench.
All experiments are based on the InternVL3-8B model \cite{zhu2025internvl3}, with detailed training parameters provided in App. \ref{trainingpara}.

\subsection{Layer-wise Adaptive Freeze-Tuning.} \label{LW‑AFT}

\subsubsection{Ablation Study.}

\begin{figure*}[ht]
  \begin{minipage}[t]{0.4\textwidth}
  \centering
    \includegraphics[height=4.1cm]{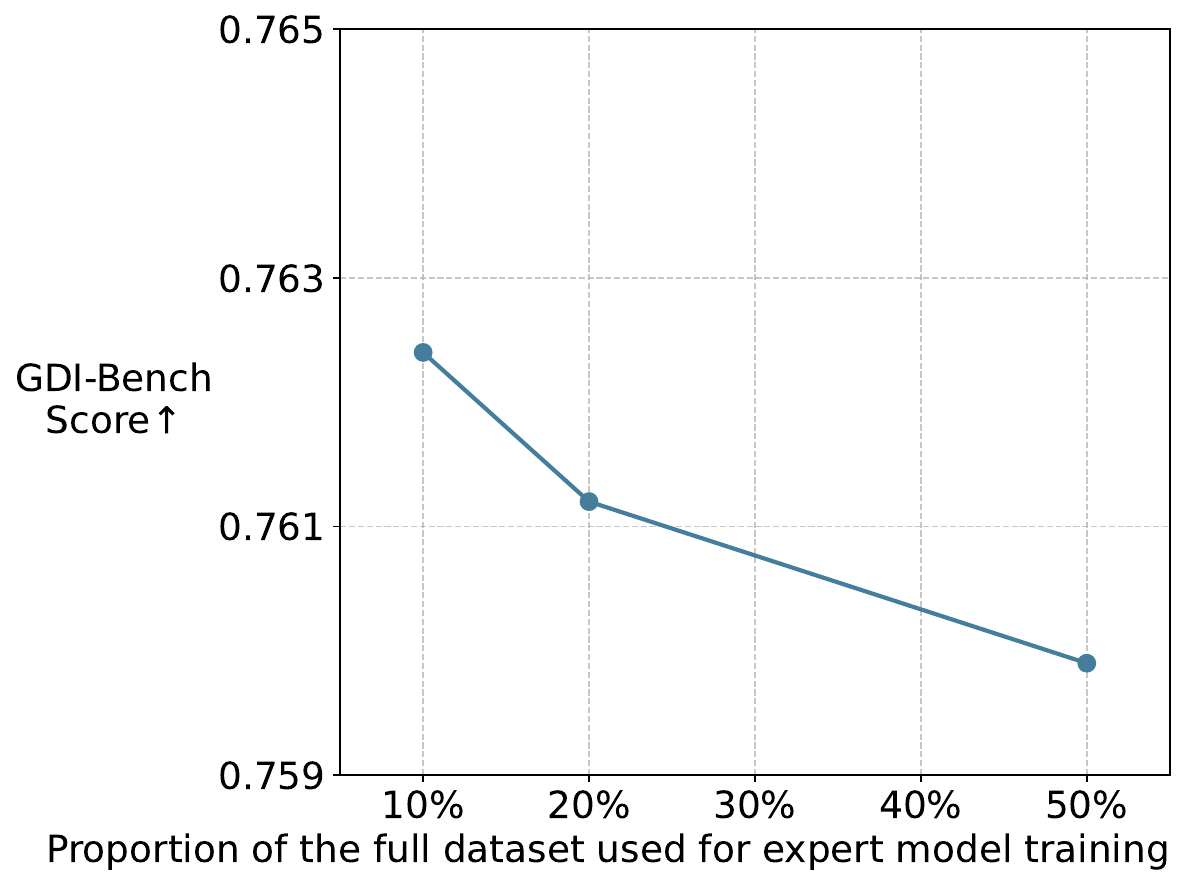}
    \caption{Ablation study on the $\alpha$.}
    \label{alpha}
  \end{minipage}
  \hfill
  \begin{minipage}[t]{0.58\textwidth}
  \centering
    \includegraphics[height=4.1cm]{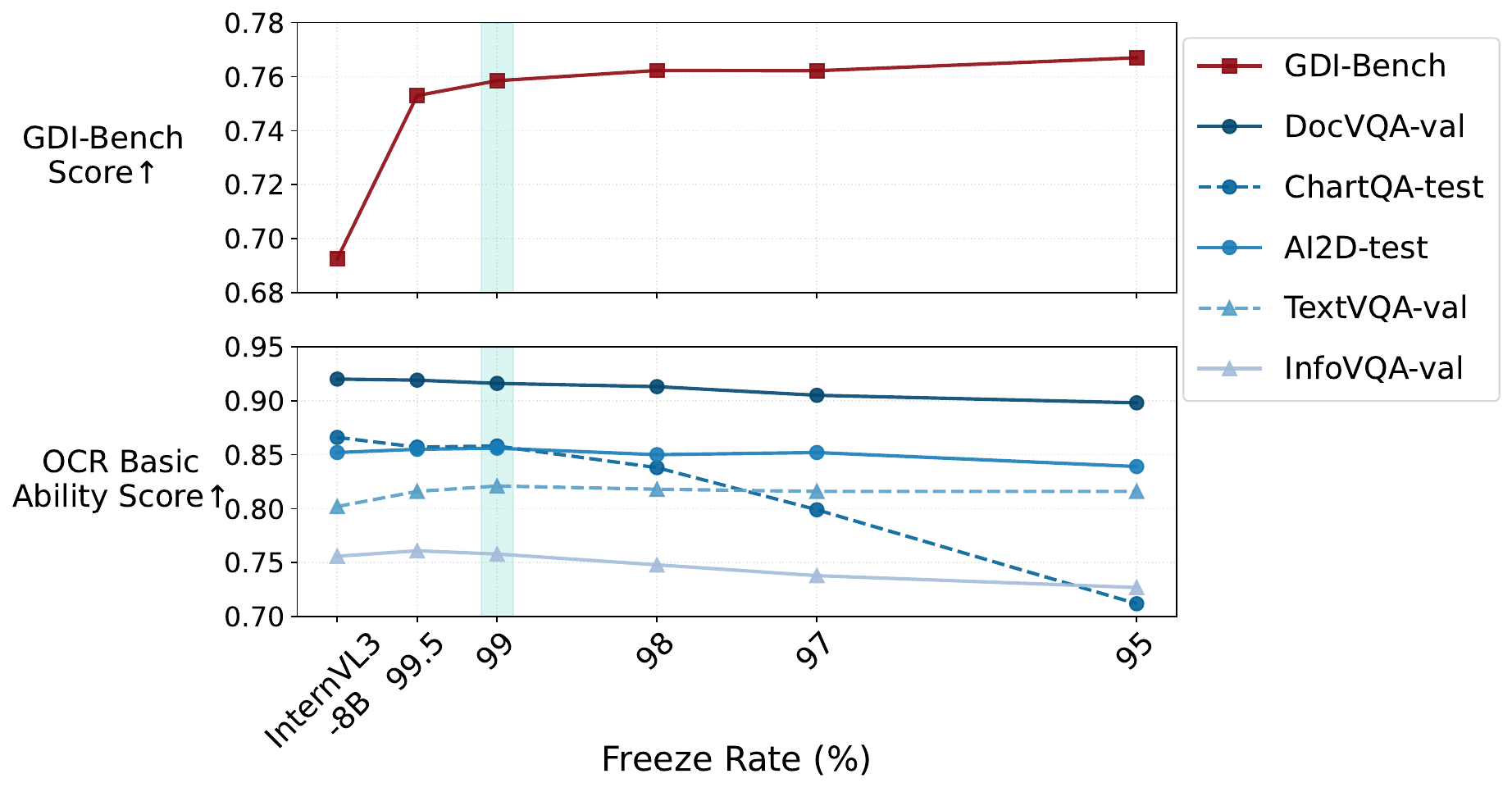}
    \caption{Ablation study on the freeze rate of parameters.}
    \label{H}
  \end{minipage}
\end{figure*}

We perform an ablation study on the parameter $\alpha$ in Section \ref{method}, which controls the fraction of the dataset used to train domain-specific expert models. 
Fig.~\ref{alpha} presents the results of this ablation study, illustrating how expert models trained with different $\alpha$ values influence the final performance on GDI-Bench. In all settings, the LW-AFT method is used with 99\% of the model parameters frozen.
The results indicate that
$\alpha=10\%$ achieves the best performance and data efficiency, and is therefore adopted in our default configuration.

We conduct an ablation study to assess the impact of varying the proportion of frozen parameters in the model. Fig.~\ref{H} illustrates the performance of the 8B-sized model across GDI-Bench and several other datasets under different freeze ratios. After evaluating the trade-off between maintaining strong performance on GDI-Bench and ensuring generalization across the additional datasets, we determine that freezing 99\% of the model parameters represents the optimal configuration for our final setup.

\subsubsection{Evaluation of Catastrophic Forgetting Severity.}
To validate the effectiveness of LW-AFT in mitigating catastrophic forgetting, we evaluate the performance of Full-Parameter Fine-Tuning, LoRA Fine-Tuning, and the LW-AFT method across a range of datasets, including OCR, Chart, Document Understanding, and Multimodal Multilingual Understanding datasets \cite{DocVQA, masry-etal-2022-chartqa, hiippala2021ai2d, TextVQA, liu2024ocrbench, ouyang2024omnidocbenchbenchmarkingdiversepdf,mathew2022infographicvqa, li2024seed2plus, liu2024mmbench}, as shown in Table \ref{general}. 
The results demonstrate that our method prevents catastrophic forgetting after SFT. Furthermore, on certain datasets, such as TextVQA, InfoVQA, and Seed2plus, our method outperforms the base model, which validates the efficacy of our intelligence-preserving training strategy.

\begin{table}[h]
\caption{The performance of different training methods on multiple datasets.}
\label{general}
\resizebox{1\linewidth}{!}{
\begin{tabular}{@{}lccccccccccc@{}}
\toprule
\multirow{2}{*}{Method}    & DocVQA$\uparrow$ & ChartQA$\uparrow$ & \multicolumn{2}{c}{AI2D$\uparrow$} & TextVQA$\uparrow$ & OCRBench$\uparrow$ & OmniDocBench$\downarrow$ & InfoVQA$\uparrow$ & Seed2plus$\uparrow$ & \multicolumn{2}{c}{MMBench$\uparrow$} \\ \cmidrule(l){2-12} 
                           & val              & test              & test             & no mask         & test              & -                  & -                        & val               & -                   & en                & cn                \\ \midrule
Base                       & 92.0             & 86.6              & 85.2             & 92.6            & 80.2              & 880                & 0.21                     & 75.6              & 69.7                & 85.5              & 85.6              \\ \midrule
Full-Parameter Fine-Tuning & 39.4             & 33.3              & 43.1             & 49.4            & 22.7              & 228                & 0.42                     & 25.9              & 44.8                & 15.6              & 14.6              \\
LoRA Fine-Tuning           & 91.3             & \textbf{85.8}     & 85.4             & 93.1            & 81.6              & 862                & \textbf{0.23}            & 74.4              & 69.7                & \textbf{85.2}     & \textbf{84.8}     \\ \midrule
LW-AFT (Ours)              & \textbf{91.6}    & \textbf{85.8}     & \textbf{85.6}    & \textbf{93.2}   & \textbf{82.1}     & \textbf{871}       & \textbf{0.23}            & \textbf{75.8}     & \textbf{70.6}       & \textbf{85.2}     & \textbf{84.8}     \\ \bottomrule
\end{tabular}}
\end{table}

\subsubsection{Cross-domain and Cross-task Evaluation.}
To validate the generalization capability of the LW-AFT method in cross-domain and cross-task scenarios, we designed the experiments covering the following two types of tasks.

Cross-domain experiments with the same task.
\begin{itemize}
    \item T1: The model is trained for the paragraph beginning extracting task in the Scientific paper domain, and evaluated on the same task in the Infograph domain.
    \item T2: The model is trained for the information extraction and format organization task in the Exam paper domains, and evaluated on the same task in the Infograph domain.
    \item T3: The model is trained for the reasoning and question-answering task for tables in the scientific paper domain, and evaluated on the same task in the Financial report domain.
\end{itemize}
Cross-task experiments with the same domain.
\begin{itemize}
    \item T4: In the Newspaper domain, the model is trained for header information extraction and format organization tasks, and evaluated on fine-grained header field extraction performance (including publication date, editor information, email address, and phone number extraction).
\end{itemize}

\begin{table}[h]
\caption{The performance of different training methods on cross-domain and cross-task scenarios.}
\label{task123}
\centering
\resizebox{0.7\linewidth}{!}{
\begin{tabular}{@{}lccccccc@{}}
\toprule
\multirow{2}{*}{Method}     & T1             & T2             & T3             & \multicolumn{4}{c}{T4}                                            \\ \cmidrule(l){2-8} 
                           &                &                &                & date           & editor         & email          & phone          \\ \midrule
Base             & 0.216          & 0.383          & 0.840          & 0.101          & 0.884          & 0.398          & 0.398          \\ \midrule
Full-Parameter Fine-Tuning & \textbf{0.040} & 0.913          & 0.600          & 0.864          & 0.992          & 0.783          & 0.989          \\
LoRA Fine-Tuning           & 0.102          & 0.473          & 0.300          & 0.093          & 0.606          & 0.171          & 0.172          \\ \midrule
LW-AFT (Ours)           & 0.096          & \textbf{0.365} & \textbf{0.200} & \textbf{0.010} & \textbf{0.316} & \textbf{0.058} & \textbf{0.153} \\ \bottomrule
\end{tabular}}
\end{table}
We compare the performance of Full-Parameter Fine-Tuning, LoRA Fine-Tuning, and the LW-AFT method under both settings, as shown in Table \ref{task123}, which displays the normalized edit distance for each task. 
As can be seen, our method demonstrates strong cross-domain and cross-task capabilities, significantly outperforming the LoRA fine-tuning, further demonstrating the effectiveness of the LW-AFT method's generalization ability in the task dimension.

\begin{table}[h]
\caption{The performance of different open-source and closed-source large models and different training methods on GDI-Bench.}
\label{gdi-bench}
\resizebox{1\linewidth}{!}{
\begin{tabular}{@{}lclllllllllc@{}}
\toprule
\multicolumn{1}{c}{\textbf{Model}} & \textbf{Size} & \multicolumn{1}{c}{\textbf{R0V0}} & \multicolumn{1}{c}{\textbf{R0V1}} & \multicolumn{1}{c}{\textbf{R0V2}} & \multicolumn{1}{c}{\textbf{R1V0}} & \multicolumn{1}{c}{\textbf{R1V1}} & \multicolumn{1}{c}{\textbf{R1V2}} & \multicolumn{1}{c}{\textbf{R2V0}} & \multicolumn{1}{c}{\textbf{R2V1}} & \multicolumn{1}{c}{\textbf{R2V2}} & \textbf{Overall} \\ \midrule
Claude-3-5-Sonnet (2024-10-22)     & 175B          & 0.90                              & 0.26                              & 0.22                              & 0.42                              & 0.63                              & 0.20                              & 0.86                              & 0.59                              & 0.44                              & 0.509            \\
Gemini-2.0-Flash                   & 750B          & 0.94                              & 0.58                              & 0.48                              & 0.37                              & 0.77                              & 0.69                              & 0.92                              & 0.73                              & 0.41                              & 0.662            \\
GPT-4o (2024-11-20)                & 200B          & 0.96                              & 0.60                              & 0.43                              & 0.39                              & 0.71                              & 0.57                              & 0.81                              & 0.51                              & 0.35                              & 0.593            \\
Grok-2-Vision-on                   & 500B          & 0.67                              & 0.20                              & 0.16                              & 0.21                              & 0.31                              & 0.30                              & 0.77                              & 0.46                              & 0.41                              & 0.377            \\
Qwen2.5-VL-72B-Instruct            & 72B           & \textbf{0.97}                     & \textbf{0.69}                     & \textbf{0.57}                     & 0.60                              & 0.81                              & 0.72                              & 0.90                              & 0.63                              & 0.49                              & 0.706            \\ \midrule
InternVL3-8B                       & 8B            & 0.96                              & 0.65                              & 0.56                              & 0.35                              & 0.77                              & 0.47                              & 0.89                              & 0.62                              & 0.50                              & 0.652            \\ \midrule
Full-Parameter Fine-Tuning Model   & 8B            & 0.94                              & 0.56                              & 0.39                              & 0.64                              & 0.71                              & 0.67                              & 0.65                              & 0.44                              & 0.51                              & 0.599            \\
LoRA Fine-Tuning Model             & 8B            & \textbf{0.97}                     & 0.66                              & 0.56                              & 0.87                              & \textbf{0.82}                     & \textbf{0.77}                     & 0.91                              & 0.71                              & \textbf{0.68}                     & 0.759            \\ \midrule
GDI-Model (Ours)                   & 8B            & \textbf{0.97}                     & 0.65                              & 0.56                              & \textbf{0.89}                     & \textbf{0.82}                     & \textbf{0.77}                     & \textbf{0.93}                     & \textbf{0.74}                     & \textbf{0.68}                     & \textbf{0.762}   \\ \bottomrule
\end{tabular}}
\end{table}

\subsection{Benchmark Evaluation Results.}\label{GDIbench}
We employ the LW-AFT method on the complete training dataset, resulting in the development of an 8B-sized GDI-Model.
To assess the effectiveness of our GDI-Model on generalized OCR tasks, we compare it against several SOTA vision-language models—namely Qwen2.5VL-72B \cite{Qwen-VL}, Gemini-2.0-Flash \cite{team2023gemini}, GPT-4o (2024-11-20) \cite{hurst2024gpt}, Claude-3-5-Sonnet, and  Grok-2-Vision-on and InternVL3-8B \cite{zhu2025internvl3} on GDI-Bench’s multi-level challenge suite (V0-R0 through V2-R2). All evaluations were conducted with the same preprocessing pipeline, as shown in Fig.~\ref{gdi-bench-eval} and Table~\ref{gdi-bench}.

Compared to the base InternVL3-8B, our model maintains strong R0-level performance and shows clear gains at R1 and R2, validating the effectiveness of the LW-AFT method. GDI-Model matches Qwen2.5-VL-72B at R0 and outperforms it at higher reasoning levels. 
At R2, our model rivals Gemini-2.0-Flash under low to medium visual complexity (V0/V1), and even surpasses it under high complexity (V2). 
These results highlight that, despite a smaller parameter size, our model achieves or exceeds the performance of much larger open- and closed-source counterparts.

\section{Conclusion}

We introduce GDI-Bench, a document-domain benchmark with broad coverage and a pioneering difficulty grading system. By decoupling visual and reasoning complexity, it enables systematic evaluation and model optimization guidance.
Alongside, we introduce LW-AFT (Layer-wise Adaptive Freeze-Tuning) and the corresponding GDI-Model, which reduces catastrophic forgetting during SFT. This method preserves general capabilities while boosting cross-domain and cross-task performance.
GDI-Model achieves strong results on GDI-Bench and other benchmarks.
We hope that the GDI-Bench can help the advancement of future models and perhaps inspire new theoretical insights.

% \section{Acknowledgement}
% The research was supported by Shanghai Artificial Intelligence Laboratory.

{
\small
\bibliographystyle{unsrt}

\bibliography{reflist}
}

%%%%%%%%%%%%%%%%%%%%%%%%%%%%%%%%%%%%%%%%%%%%%%%%%%%%%%%%%%%%
\newpage
\appendix

\section{Technical Appendices and Supplementary Material}
\subsection{Training Details and Hyperparameters}\label{trainingpara}

We use the InternVL3-8B model as the base model and perform supervised fine-tuning on it. In our experiments, both the LW-AFT method and Full-Parameter Fine-Tuning utilize the same set of training hyperparameters, as shown in Table \ref{traininghyperpara}. The training process was conducted on 8 A100 GPUs.
The specific hyperparameters for LoRA fine-tuning are provided in Table  \ref{traininghyperparalora}.

\begin{table}[h]
\centering
\caption{Hyperparameters of the Full-Parameter Fine-Tuning and Layer-wise Adaptive Freeze-Tuning.}\label{traininghyperpara}
\begin{tabular}{@{}ll@{}}
\toprule
Hyperparameters   & Value  \\ \midrule
Force image size  & 448    \\
Drop path rate    & 0.1    \\
Max dynamic patch & 20     \\
Train epochs      & 1      \\
Learning rate     & 3e-5   \\
Weight decay      & 0.05   \\
Warm-up ratio     & 0.03   \\
Lr scheduler      & cosine \\
Batch size        & 12     \\
Max seq length    & 8192   \\ \bottomrule
\end{tabular}
\end{table}

\begin{table}[h]
\centering
\caption{Hyperparameters of the LoRA Fine-Tuning.}\label{traininghyperparalora}
\begin{tabular}{@{}ll@{}}
\toprule
Hyperparameters   & Value  \\ \midrule
Force image size  & 448    \\
Drop path rate    & 0.0    \\
Use LLM LoRA      & 16     \\
Max dynamic patch & 20     \\
Train epochs      & 1      \\
Learning rate     & 3e-5   \\
Weight decay      & 0.05   \\
Warm-up ratio     & 0.03   \\
Lr scheduler      & cosine \\
Batch size        & 12     \\
Max seq length    & 8192   \\ \bottomrule
\end{tabular}
\end{table}

\subsection{Training Data}\label{trainingset}

We construct a training dataset for supervised fine-tuning to enhance the base model InternVL3-8B's capabilities in the R1 and R2 domains. This dataset contains approximately 200,000 QA-form tasks, primarily focused on tasks similar to R1 and R2, but not identical. The data sources are strictly different from those of GDI-Bench. Detailed information on the dataset size and task settings can be found in Table \ref{trainingset20w}.

\begin{table}[t]
\caption{Composition of the training set.}\label{trainingset20w}
\resizebox{1\linewidth}{!}{
\begin{tabular}{@{}llc@{}}
\toprule
Domain              & Task                                                                                                                                                                                                                                                          & Training data \\ \midrule
Newspaper           & \begin{tabular}[c]{@{}l@{}}Extract header information of a newspaper page. \\ Answer reasoning questions for newspaper page (/ multiple choice).……\end{tabular}                                                                                                                    & 29k           \\ \midrule
Scientific Paper    & \begin{tabular}[c]{@{}l@{}}Extract the author information of the paper.\\ Answer reasoning questions for a paragraph on a scientific paper page (/ multiple choice).……\end{tabular} & 108k          \\ \midrule
Infograph           & \begin{tabular}[c]{@{}l@{}}Extract the maintitle.\\ Answer reasoning questions for a infographic page (/ multiple choice).……\end{tabular}                                                                                                                     & 32k           \\ \midrule
Exam paper          & \begin{tabular}[c]{@{}l@{}}Full text extraction of exampaper pages.\\ Extract the content of the exam paper based on the type of question and the question number.……\end{tabular}                                                                             & 12k           \\ \midrule
Financial report    & \begin{tabular}[c]{@{}l@{}}Extract the table on a page and convert to LaTeX format.\\ Answer reasoning questions for a table on a financial report page (/ multiple choice).……\end{tabular}                                                               & 3k            \\ \midrule
Handwritten Content & Recognize handwritten content in English and Chinese.                                                                                                                                                                                                         & 12k           \\ \bottomrule
\end{tabular}}
\end{table}

\subsection{Qualitative Analysis.}
\begin{figure}[H]
    \centering
    \includegraphics[width=1\linewidth]{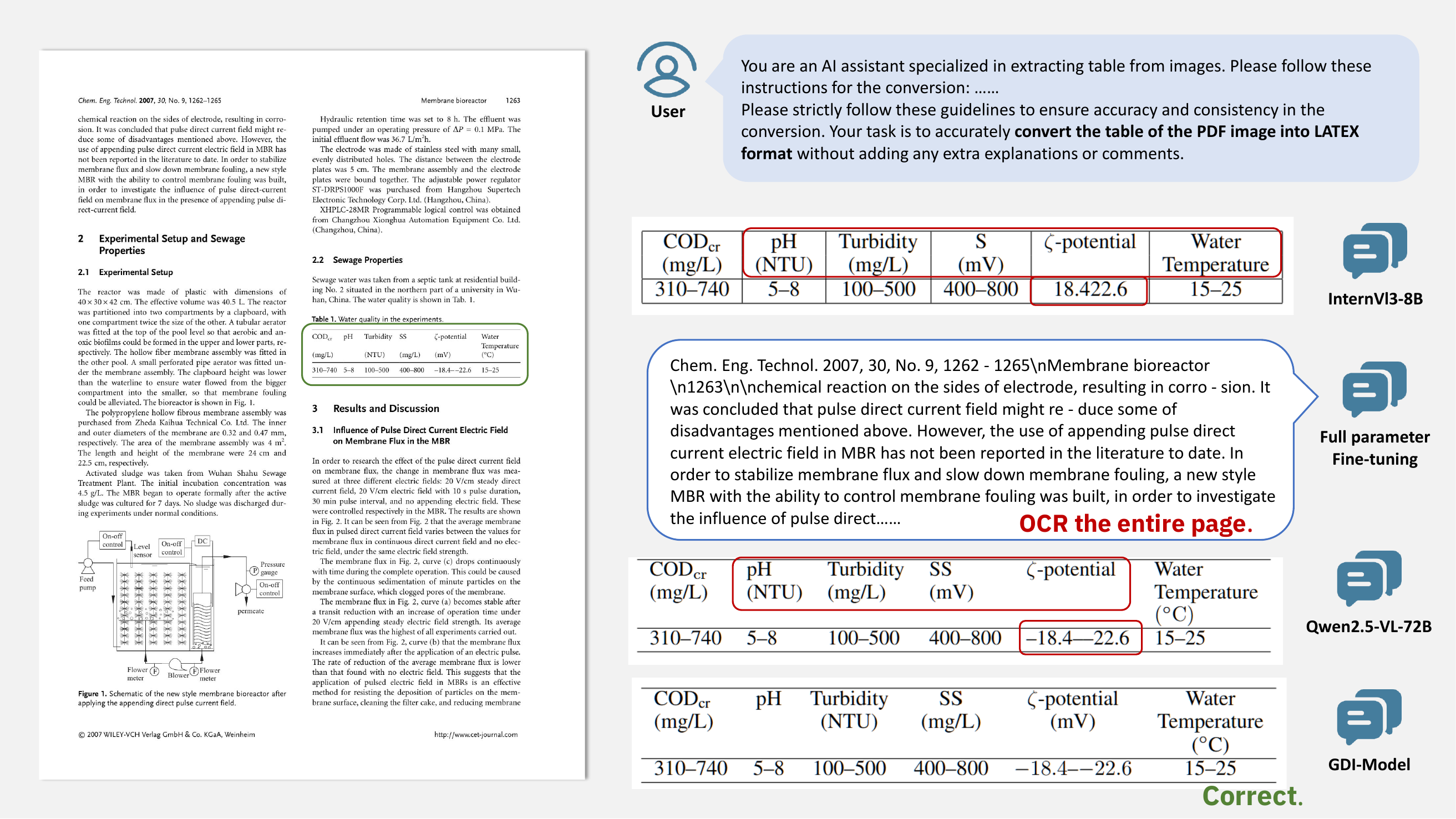}
    \caption{Table extraction task, both the InternVL3-8B model and the Qwen2.5-VL-72B model made errors. The full parameter fine-tuned models failed to handle the table extraction task due to catastrophic forgetting and ended up performing OCR on the entire page’s content. In contrast, the GDI-Model successfully extracted the table and output it in LaTeX format. }\label{case1}
\end{figure}

\begin{figure}[H]
    \centering
    \includegraphics[width=1\linewidth]{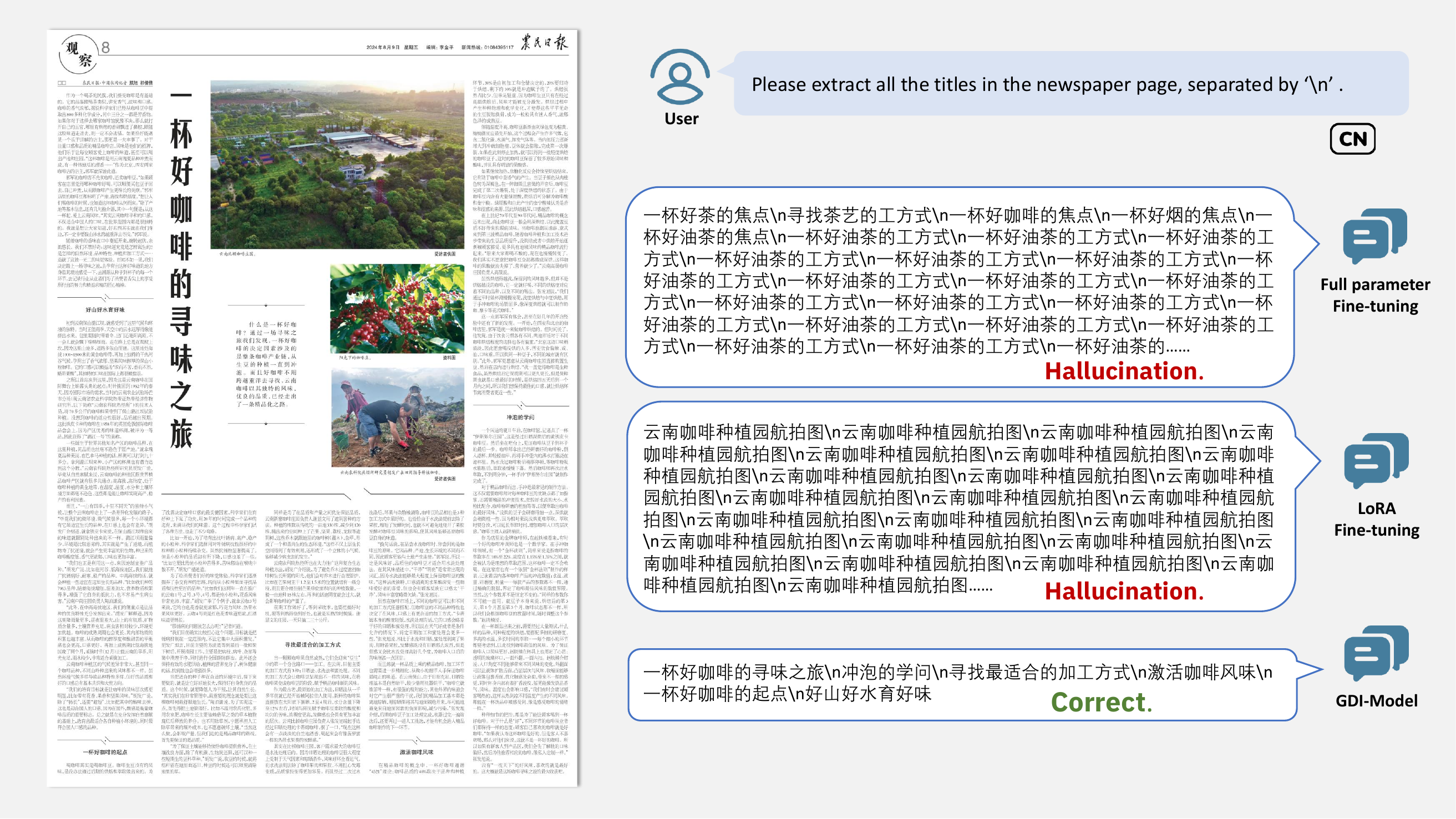}
    \caption{Newspaper titles extraction task, both full parameter fine-tuned models and LoRA fine-tuned models tend to generate model hallucinations, whereas the GDI-Model, obtained using the LW-AFT method, is able to complete the task correctly.}\label{case2}
\end{figure}

\begin{figure}[H]
    \centering
    \includegraphics[width=1\linewidth]{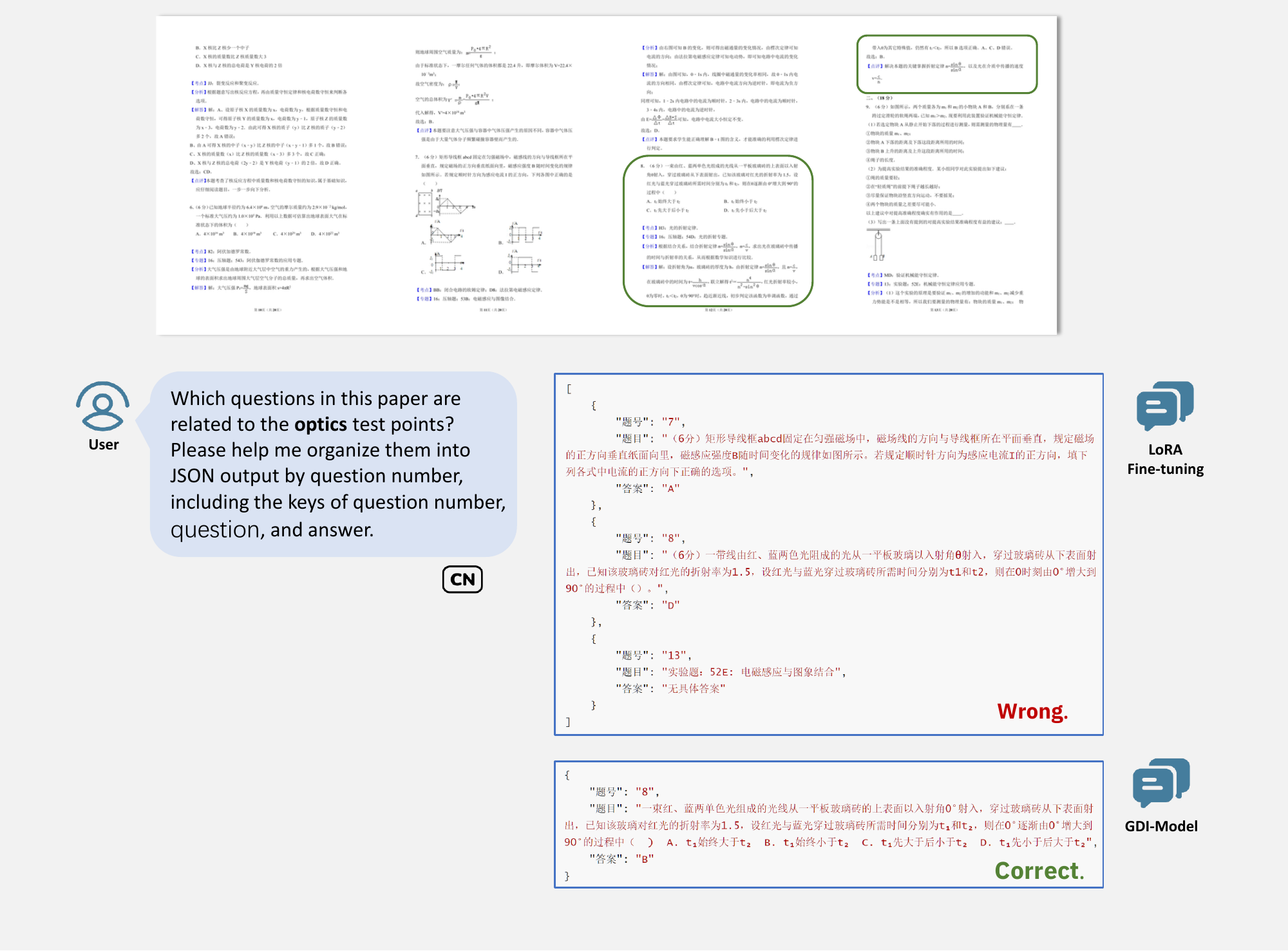}
    \caption{Task of extracting and organizing test points into JSON format, the LoRA model fails to accurately locate the key points or produce correct outputs, whereas the GDI-Model, obtained using the LW-AFT method, is able to complete the task correctly.}\label{case3}
\end{figure}

\begin{figure}[H]
    \centering
    \includegraphics[width=1\linewidth]{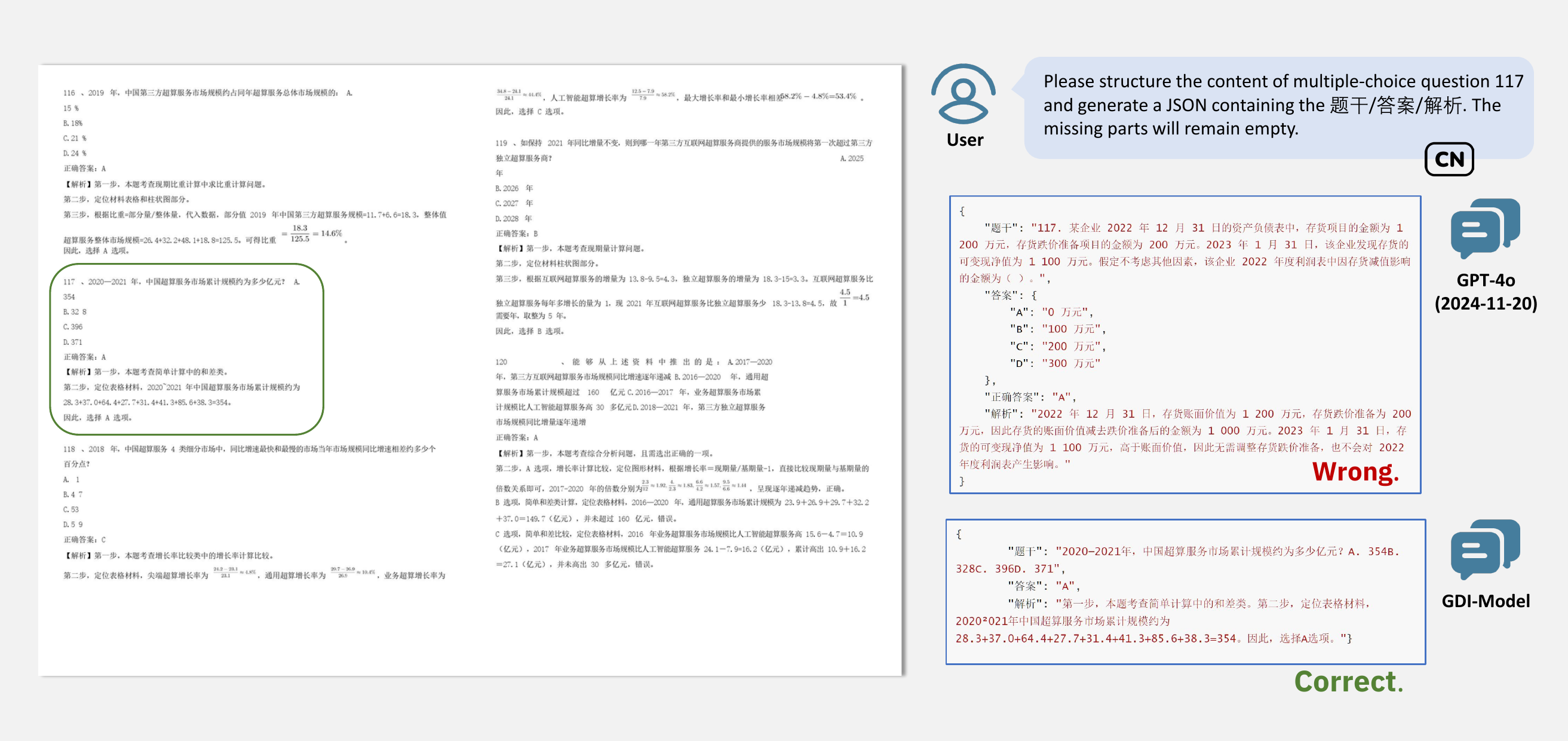}
    \caption{The task of extracting and organizing problems by question number into JSON format, the GPT-4o model fails to correctly locate the questions based on their numbers and ends up fabricating non-existent questions. In contrast, the GDI-Model successfully extracts the questions and outputs them in the specified JSON format. }\label{case4}
\end{figure}

\begin{figure}[H]
    \centering
    \includegraphics[width=1\linewidth]{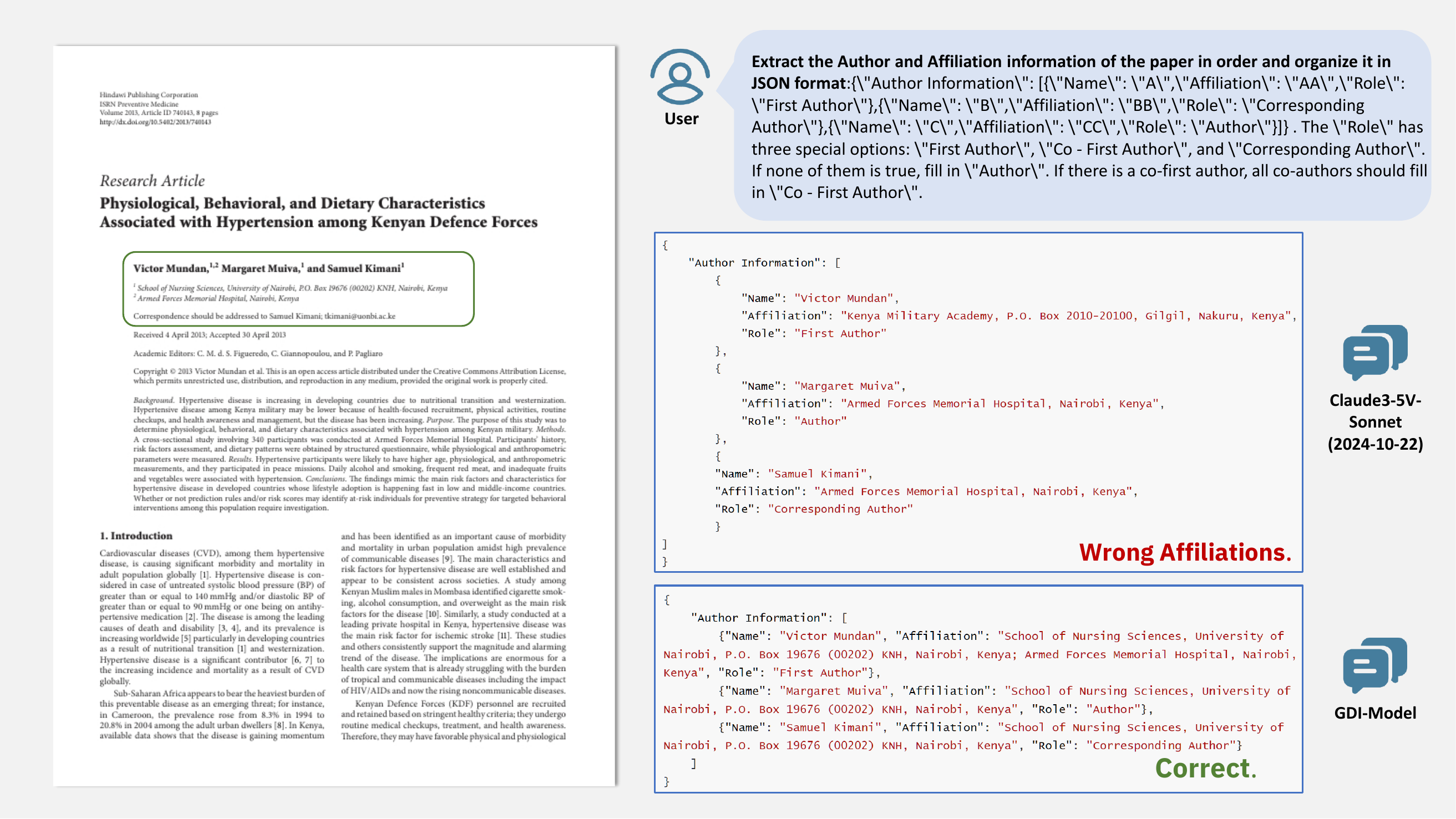}
    \caption{The task of extracting paper author information and organizing it into JSON format, the Claude3-5V-Sonnet model extracted incorrect affiliation information, while the GDI-Model successfully completed the task.}\label{case5}
\end{figure}

\begin{figure}[H]
    \centering
    \includegraphics[width=1\linewidth]{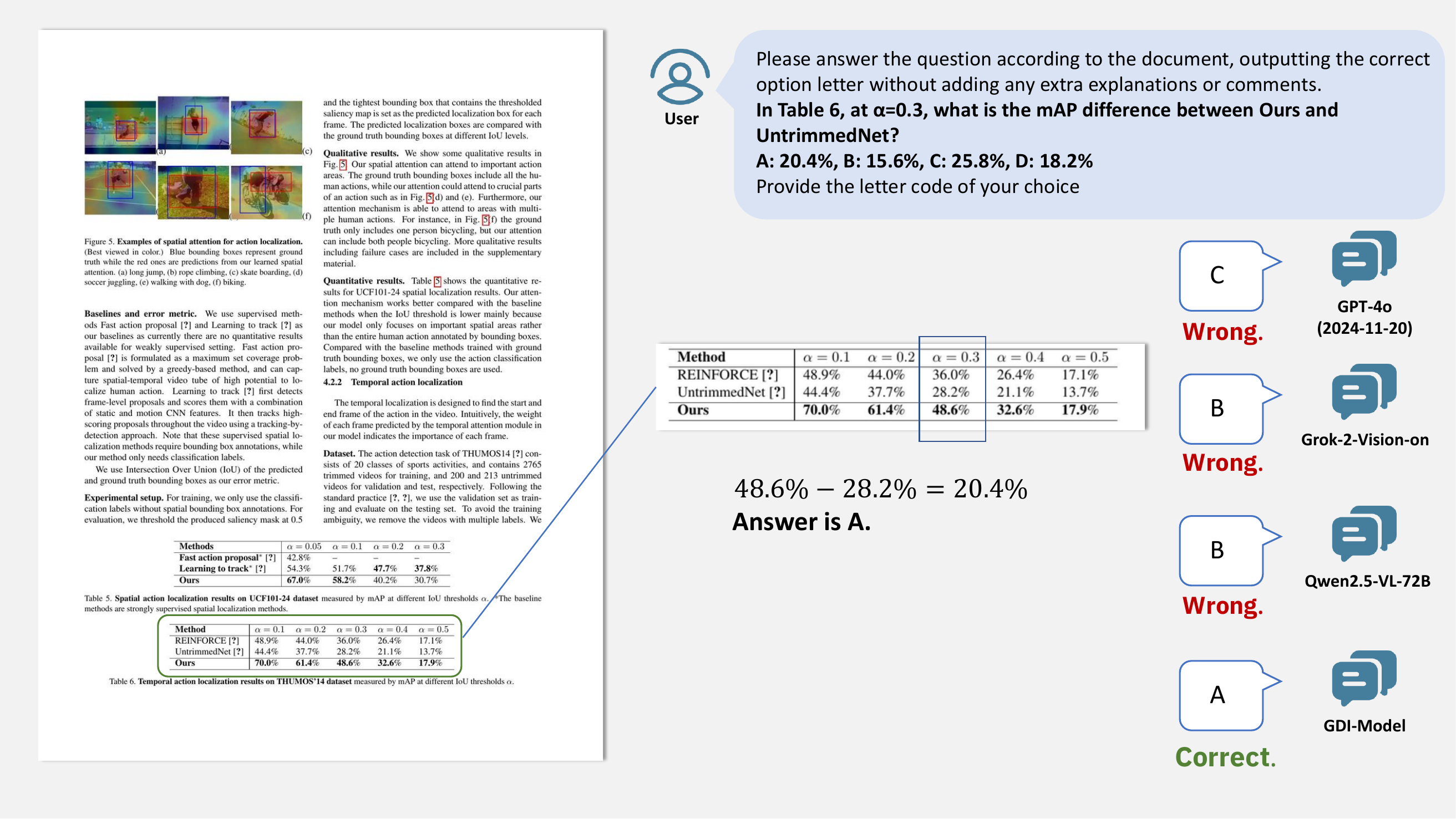}
    \caption{Table reasoning tasks in scientific paper page, the GPT-4o, Grok-2-Vision-on, and Qwen2.5-VL-72B models provided incorrect answers, whereas the GDI-Model answered correctly.}\label{case6}
\end{figure}

\begin{figure}[H]
    \centering
    \includegraphics[width=1\linewidth]{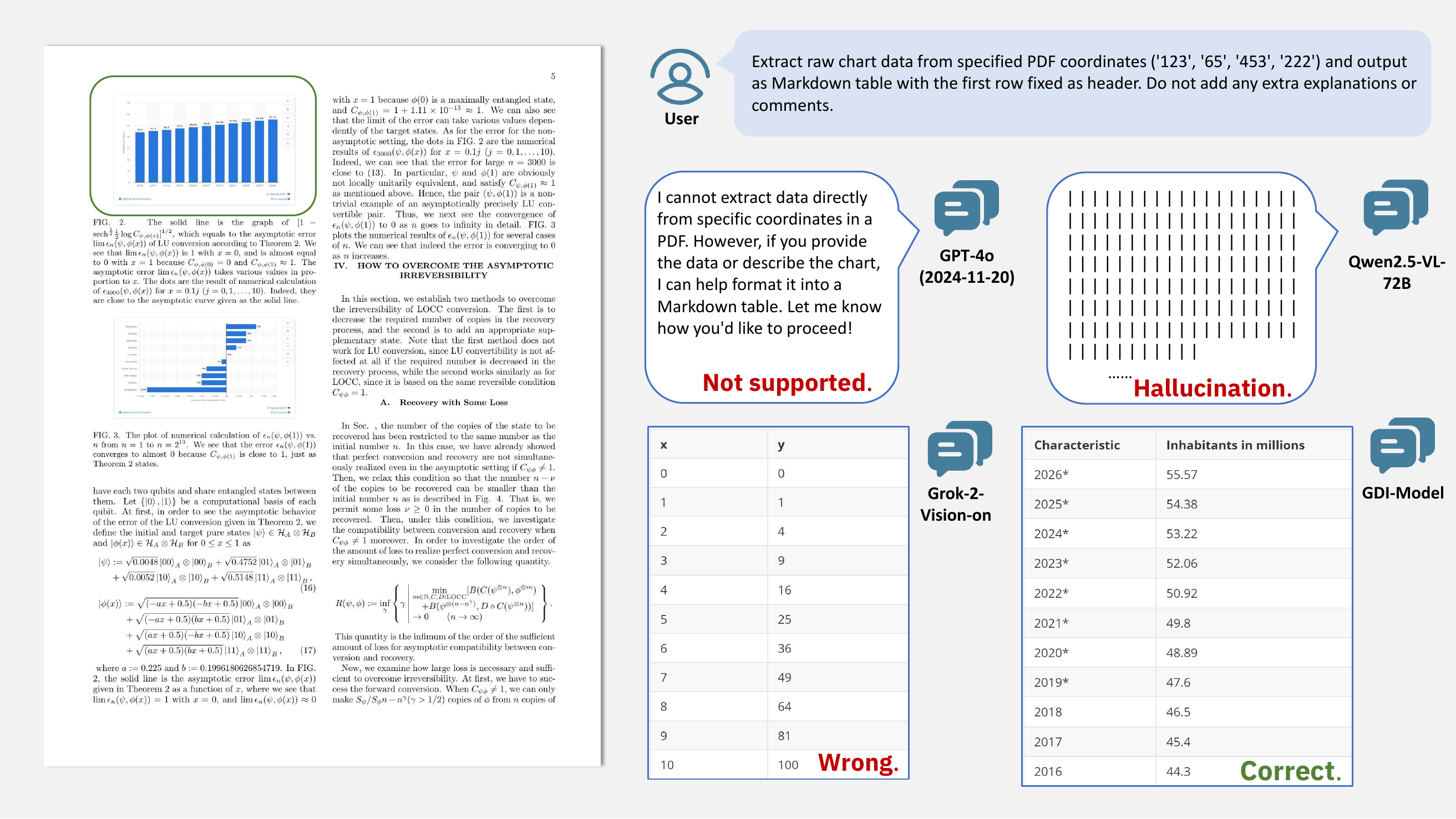}
    \caption{The task of converting chart data within a region into a markdown table, the GPT-4o model does not support the task, Qwen2.5-VL-72B produced a model hallucination, and Grok-2-Vision-on incorrectly converted the data and failed to extract it. In contrast, the GDI-Model completed the task.}\label{case7}
\end{figure}

In this section, we provide a qualitative analysis, comparing several cases to demonstrate the performance of the GDI model, as shown in Figures \ref{case1}-\ref{case7}.

\subsection{Sample Prompts for Evaluation.}

\begin{tcolorbox}[boxrule=1pt, boxsep=2pt, colback=gray!20, fontupper=\normalsize, fonttitle=\scriptsize, title=Prompt template for document extraction (v0 tasks)]
Extract all text from the image accurately without interpretation and output only the text without adding any extra explanations or comments.
\end{tcolorbox}

\begin{tcolorbox}[boxrule=1pt, boxsep=2pt, colback=gray!20, fontupper=\normalsize, fonttitle=\scriptsize, title=Prompt template for document conversion (v1 \& v2 tasks)]
\parbox{\linewidth}{
You are an AI assistant specialized in converting PDF images to Markdown format. Please follow these instructions for the conversion:
\begin{itemize}
  \item \textbf{1. Text Processing:}
  \begin{itemize}
    \item Accurately recognize all text content in the PDF image without guessing or inferring.
    \item Convert the recognized text into Markdown format.
    \item Maintain the original document structure, including headings, paragraphs, lists, etc.
  \end{itemize}
  \item \textbf{2. Mathematical Formula Processing:}
  \begin{itemize}
    \item Convert all mathematical formulas to LaTeX format.
    \item Enclose inline formulas with \texttt{\textbackslash( \textbackslash)}. For example: This is an inline formula \( E = mc^2 \).
    \item Enclose block formulas with \texttt{\textbackslash[ \textbackslash]}
. For example: \[ \frac{-b \pm \sqrt{b^2 - 4ac}}{2a} \]
  \end{itemize}
  \item \textbf{3. Table Processing:}
  \begin{itemize}
    \item Convert tables to Markdown format.
  \end{itemize}
  \item \textbf{4. Figure Handling:}
  \begin{itemize}
    \item Ignore figures content in the PDF image. Do not attempt to describe or convert images.
  \end{itemize}
  \item \textbf{5. Output Format:}
  \begin{itemize}
    \item Ensure the output Markdown document has a clear structure with appropriate line breaks between elements.
    \item For complex layouts, try to maintain the original document's structure and format as closely as possible.
  \end{itemize}
\end{itemize}
Please strictly follow these guidelines to ensure accuracy and consistency in the conversion. Your task is to accurately convert the content of the PDF image into Markdown format without adding any extra explanations or comments.
}
\end{tcolorbox}

\begin{tcolorbox}[boxrule=1pt, boxsep=2pt, colback=gray!20, fontupper=\normalsize, fonttitle=\normalsize, title=Prompt template for newspaper date extraction task]
Extract the date information (only including the day, month, and year, if available) of the newspaper, preserving the original punctuation and formatting.
\end{tcolorbox}

\begin{tcolorbox}[boxrule=1pt, boxsep=2pt, colback=gray!20, fontupper=\normalsize, fonttitle=\normalsize, title=Prompt template for extracting tables from documents]
You are an AI assistant specialized in extracting table from images. Please follow these instructions for the conversion:
\begin{itemize}
  \item \textbf{1. Text Processing:}
  \begin{itemize}
    \item Accurately recognize text content in the table without guessing or inferring.
  \end{itemize}
  \item \textbf{2. Mathematical Formula Processing:}
  \begin{itemize}
    \item Convert all mathematical formulas to LaTeX format.
    \item Enclose inline formulas with \verb|\(| \verb|\)|. For example: This is an inline formula \( E = mc^2 \).
    \item Enclose block formulas with \verb|\[| \verb|\]|. For example: \[ \frac{-b \pm \sqrt{b^2 - 4ac}}{2a} \].
  \end{itemize}
  \item \textbf{3. Table Processing:}
  \begin{itemize}
    \item Convert tables to LATEX format.
    \item Start with \verb|\begin{tabular}|.
  \end{itemize}
  \item \textbf{4. Output Format:}
  \begin{itemize}
    \item Ensure the output LATEX has a clear structure with appropriate line breaks between elements.
    \item For complex layouts, try to maintain the original document's structure and format as closely as possible.
  \end{itemize}
\end{itemize}
Please strictly follow these guidelines to ensure accuracy and consistency in the conversion. Your task is to accurately convert the table of the PDF image into LATEX format without adding any extra explanations or comments.
\end{tcolorbox}

\begin{tcolorbox}[boxrule=1pt, boxsep=2pt, colback=gray!20, fontupper=\normalsize, fonttitle=\normalsize, title=Prompt template for single-choice questions]
Please answer the question according to the document, outputting the correct option letter without adding any extra explanations or comments.
\end{tcolorbox}

\begin{tcolorbox}[boxrule=1pt, boxsep=2pt, colback=gray!20, fontupper=\normalsize, fonttitle=\normalsize, title=Prompt template for chart extraction from document]
Extract raw chart data from specified PDF coordinates (x1, y1, x2, y2) and output as Markdown table with the first row fixed as header. Do not add any extra explanations or comments.
\end{tcolorbox}

\begin{tcolorbox}[boxrule=1pt, boxsep=2pt, colback=gray!20, fontupper=\normalsize, fonttitle=\normalsize, title=Prompt template for targeted extraction]
Please extract the text in the yellow color box in the document without adding any extra explanations or comments.
\end{tcolorbox}

\subsection{Limitations} \label{lim}
In this section, we outline the limitations of GDI-Bench. Currently, GDI-Bench supports only single-image document understanding tasks, excluding multi-image or multi-document tasks. Future work will expand the benchmark to include these more complex tasks, enhancing its difficulty and enabling wider evaluations.

\subsection{Broader impacts} \label{boradimpact}
GDI-Bench helps developers diagnose model failures in vision-reasoning coupled document understanding tasks, aiding in weakness identification and optimization. However, the decoupling analysis method could be exploited maliciously, potentially boosting the capabilities of MLLMs (Multi-modal Large Language Models) in harmful ways.

\end{document}